\documentclass[conference]{IEEEtran}
\IEEEoverridecommandlockouts
\usepackage{cite}
\usepackage{amsmath,amssymb,amsfonts}
\usepackage{algorithmic}
\usepackage{graphicx}
\usepackage{textcomp}
\usepackage[dvipsnames]{xcolor}
\usepackage{enumitem}
\def\BibTeX{{\rm B\kern-.05em{\sc i\kern-.025em b}\kern-.08em
    T\kern-.1667em\lower.7ex\hbox{E}\kern-.125emX}}

\usepackage{url}
\usepackage{comment}

\begin{document}

\title{Scalable Interactive Machine Learning for Future Command and Control
}

\author{Anna Madison$^{*1}$, Ellen Novoseller$^{*1}$, Vinicius G. Goecks$^{*1}$, Benjamin T. Files$^{1}$, Nicholas Waytowich$^{1}$, \\ Alfred Yu$^{1}$, Vernon J. Lawhern$^{1}$, Steven Thurman$^{1}$, Christopher Kelshaw$^{2}$, and Kaleb McDowell$^{1}$ \\ \\
\IEEEauthorblockN{$^1$ Humans in Complex Systems, U.S. DEVCOM Army Research Laboratory, \\ Aberdeen Proving Ground, MD, USA}
\IEEEauthorblockN{$^2$ U.S. Mission Command Battle Lab, Futures Branch, Ft. Leavenworth, KS, USA} \\ 
\IEEEauthorblockN{Correspondence: Anna Madison, anna.m.madison2.civ@army.mil  \vspace{-1em}}
\thanks{*These three authors contributed equally to this work.}
\thanks{This research was sponsored by the Army Research Laboratory and accomplished under Cooperative Agreement \#W911NF-23-2-0072.} 
\thanks{This paper was originally presented at the NATO Science and Technology Organization Symposium (ICMCIS) organized by the Information Systems Technology (IST) Panel, IST-205-RSY – the ICMCIS, held in Koblenz, Germany, 23-24 April 2024.}
}

\maketitle

\begin{abstract}
Future warfare will require Command and Control (C2) personnel to make decisions at shrinking timescales in complex and potentially ill-defined situations. Given the need for robust decision-making processes and decision-support tools, integration of artificial and human intelligence holds the potential to revolutionize the C2 operations process to ensure adaptability and efficiency in rapidly changing operational environments. We propose to leverage recent promising breakthroughs in interactive machine learning, in which humans can cooperate with machine learning algorithms to guide machine learning algorithm behavior. This paper identifies several gaps in state-of-the-art science and technology that future work should address to extend these approaches to function in complex C2 contexts. In particular, we describe three research focus areas that together, aim to enable scalable interactive machine learning (SIML): 1) developing human-AI interaction algorithms to enable planning in complex, dynamic situations; 2) fostering resilient human-AI teams through optimizing roles, configurations, and trust; and 3) scaling algorithms and human-AI teams for flexibility across a range of potential contexts and situations. 
\end{abstract}


\section{Introduction}

Future warfare will place unprecedented and dramatically-increased demands on Command and Control (C2)\footnote{The first  ``C"  in C2, command, is the authoritative act of making decisions and ordering action, with key elements being authority, responsibility, decision-making, and leadership. The second ``C" in C2, control, is defined as the act of monitoring and influencing command action through direction, feedback, information, and communications \cite{army2022FM6-0}.} systems.\footnote{C2 systems specify arrangements of people, processes, networks, and command posts~\cite{army2019ADP6-0}.}
Maintaining decision advantage over adversaries during multi-domain operations (MDO) will require robust C2 decision-making at shorter timescales in more complex and potentially ill-defined situations. Dispersed, isolated, and mobile command post nodes, forces, and autonomous agents will need to maintain unity of effort in the face of Denied, Degraded, Intermittent, and Limited (DDIL) communications, while integrating complex data streams into synchronized decision-making capabilities. Yet, current C2 processes are time-consuming and consist of linear sequences of steps, and thus may not adapt well to future battlefield challenges.

To achieve decision dominance on the future battlefield, C2 systems will need to leverage recent breakthroughs in interactive machine learning\footnote{Methods for humans to cooperate with machine learning algorithms to guide and adapt their behavior toward human-desired intent, e.g., via instructions, demonstrations, or in-the-loop feedback.} and hybrid human-machine intelligence that indicate the potential for humans and AI to work together to leverage their respective strengths~\cite{case2018become, scharre2016centaur, zhang2019leveraging}. There is, however, no one-size-fits-all approach to human-machine integration.  Metcalfe, Perelman et al.~\cite{metcalfe2021systemic} introduce a \textit{landscape of human-AI partnership}, which posits that the nature of human-AI interaction depends on task complexity, decision timescales, and information certainty. While some tasks may best be solved by AI alone (e.g., humans may not need AI to solve high-certainty, low-complexity, long-timescale tasks), typically, humans and AI agents enjoy complementary, synergistic strengths. For example, AI might efficiently traverse large datasets and explore many possible outcomes, while humans can apply task intuition and common sense and resolve ethical dilemmas, and together, humans and AI may exhibit creativity along different, complementary axes.


\begin{figure*}[ht!]
    \centerline{\includegraphics[width=1.5\columnwidth]{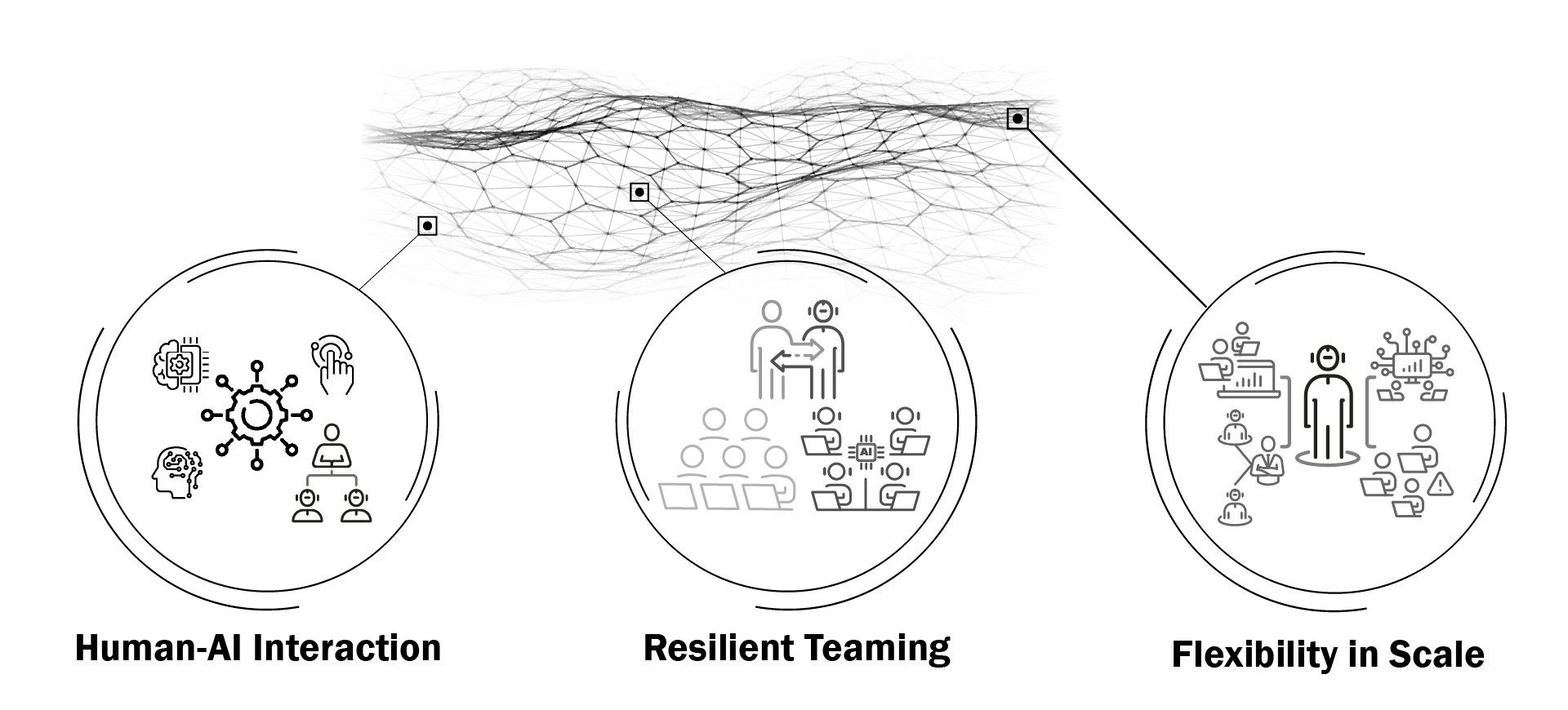}}
    \caption{\textbf{Scalable Interactive Machine Learning (SIML) research focus areas.} We propose three research focus areas in SIML to achieve the envisioned C2 developments: 1) developing human-AI interaction algorithms for planning in complex and dynamic environments; 2) fostering resilient human-AI teaming, for instance optimizing team configurations and calibrated trust; and 3) scaling methods developed in the first two research areas to succeed across anticipated future battlefield scenarios.}
    \label{fig:SC3_NATO_Graphic_smaller.png}
\end{figure*}

AI-enabled decision support systems hold the potential to revolutionize the C2 operations process\footnote{The \textit{C2 operations process} is the Army's framework for organizing and putting C2 into action, and consists of the four major C2 activities performed during operations: planning, preparing, executing, and continuously assessing~\cite{army2019ADP5-0}.} to facilitate more robust, efficient, and effective C2 in future operational environments. 
We propose that research in \textit{Scalable Interactive Machine Learning} (SIML) is critical towards enabling integrated human-AI interaction-based systems that scale to the demands of future C2 (discussed in McDowell et al.~\cite{mcdowell2024reenvisioning}, ICMCIS 2024 companion paper). 
First, SIML is critical to streamlining elements of the C2 operations process, for instance to perform rapid development and analysis of courses of action (COAs), interactively iterating on proposed alternatives based on feedback from C2 personnel; such systems could compress the military decision-making process (MDMP)\footnote{In the planning phase of the operations process, the MDMP defines the series of steps used to produce a plan or order~\cite{army2022FM6-0}.} or rapidly adjust a COA in real-time based on continuously-assessed battlefield conditions.
Second, SIML is critical to maintaining coordination and unity of effort among multiple echelons and across dispersed physical and virtual command post nodes, forces, and AI agents; e.g., SIML models would enhance the ability of dispersed units under DDIL conditions to predict each others' actions in response to unforeseen events.
Third, SIML systems 
could enable AI-based C2 planning to continually adapt based on accumulated data and interactive human feedback, retain knowledge and expertise following personnel rotation, and assist C2 personnel in making optimized decisions based on the most recent and relevant data.

To actualize the envisioned C2 operational developments, this paper proposes three research focus areas in SIML (see Figure 1), covered in the following sections: a) developing interactive machine learning algorithms for robust planning in complex and dynamic environments, b) creating effective and resilient human-AI teams, and c) scaling the approaches in a) and b) to succeed in across a wide range of possible battlefield scenarios.
The final discussion also includes practical considerations for operationalizing the envisioned processes.

\section{Planning in Complex and Dynamic Environments with Interactive Machine Learning}

The first research focus area addresses the challenges of jointly leveraging the strengths of both humans and AI to act in complex and dynamic environments. This directly applies to developing well-performing COAs for uncontrolled, uncertain environments under further challenges such as time pressure, DDIL communications, and unknown opponent intentions.

\subsection{Methods for Humans to Guide Learning and Planning Processes via a Range of Interaction Modalities} Streamlining the MDMP to simultaneously develop and analyze COAs would enable rapid generation and evaluation of many tactical and operational options~\cite{farmer2022four, schadd2022machine}, and would enable adaptation of the MDMP based on commander preferences (note that Section~\ref{ssec:multi_human_AI} discusses integrating input across multiple people).
To develop human-AI interaction methods for such tasks, new research should build on human-guided machine learning techniques that leverage human feedback to train and adapt AI algorithms based on human intentions. Such algorithms leverage humans in a variety of ways, for instance to create knowledge bases (e.g., via instruction manuals or Internet text~\cite{openai2022introducing}), to provide instructions (e.g., via language~\cite{schadd2022machine} or demonstrations of desired behavior~\cite{argall2009survey,goecks2019efficiently}), to give corrections (e.g., allowing users to adjust AI-intended actions~\cite{sharma2022correcting, goecks2023disasterresponsegpt}), or to provide performance evaluation feedback (e.g., via ratings or relative comparisons identifying user-preferred outcomes~\cite{christiano2017deep, lee2021pebble}). User input could be upfront, as is typical with instructions or demonstrations, or interactively provided during training, as can be more common with corrective or evaluative feedback. 
Notably, language-based human-AI brainstorming is a promising direction, having been used for text summarization~\cite{stiennon2020learning}; to train large language models (LLMs) such as ChatGPT, for which human labelers ranked the model's responses from worst to best~\cite{openai2022introducing}; and to propose COAs for disaster response scenarios, allowing users to collaborate with an LLM to iterate on plans~\cite{goecks2023disasterresponsegpt}.

Open research problems include how to seamlessly combine different human feedback modalities within a single learning system and how to leverage human feedback to infer human intentions for sophisticated, long-horizon planning tasks such as COA development. These advances would enable C2 personnel to interact with a learning system to adapt proposed COAs and to choose from a variety of interaction types based on situational needs, from demonstrating corrections to catching risky behaviors to communicating shifts in mission objectives to placing constraints that prohibit undesirable actions.

\subsection{Methods for Multiple Humans and Multiple AI Agents to Interact in Complex and Dynamic Environments}\label{ssec:multi_human_AI}
In the machine learning research community, several methods have been proposed for decentralized planning with multiple AI agents~\cite{matta2019q, zhang2021model, wang2021model, elleuch2021novel}; yet, there has been relatively little research on methods for interactive human-guided learning that extend to multi-agent systems or that integrate feedback from many dispersed humans.
Relevant to C2, interactive AI methods that plan over multi-agent systems would enable a COA development and analysis system to plan over many distributed forces and assets, which could be either humans or AI agents, while receiving feedback from C2 personnel to improve the proposed operational and tactical recommendations. Meanwhile, interactive AI methods that receive feedback from groups of humans would enable the system to intelligently integrate input from a range of C2 personnel, potentially with different functions, areas of expertise, biases, and mindsets. Such systems could weight feedback from various humans, e.g., depending on their expertise or input quality, and would assist with smooth transitions following personnel rotation. These methods for multi-human multi-AI interaction will lay the foundation for resiliency to DDIL conditions and to other interruptions in human availability, discussed further below.

Existing learning methods for multi-agent systems typically either optimize behavior relative to a pre-specified numerical objective function~\cite{vinyals2019grandmaster, wu2021too, matta2019q, meta2022human} or leverage upfront human demonstrations~\cite{
le2017coordinated, song2018multi, yu2019multi, meta2022human}, rather than enabling interactive human feedback throughout learning. Further research is needed to develop methods for humans to intuitively guide large groups of interacting agents, e.g., via paradigms such as multi-agent reinforcement learning (RL).

Most existing human-guided learning methods either learn from a single human or treat feedback from multiple humans as if it came from a single person~\cite{ouyang2022training, casper2023open}.
Several works introduce methods for crowdsourcing human feedback, that is, obtaining human feedback from an ensemble of humans with different biases and reliability levels. 
These approaches query humans for discrete classification labels~\cite{parisi2014ranking, sakata2019crownn, zhang2014spectral, shah2020permutation} or pairwise comparisons~\cite{zhang2022batch, chhan2024crowd, chakraborty2024maxmin, siththaranjan2023distributional} or obtain task demonstrations from multiple people~\cite{beliaev2022imitation}. Yet, these approaches often assume that all human data is provided upfront, rather than given interactively~\cite{parisi2014ranking, sakata2019crownn, zhang2014spectral, shah2020permutation, beliaev2022imitation, zhang2022batch, chakraborty2024maxmin, siththaranjan2023distributional}, do not consider sequential decision-making tasks~\cite{parisi2014ranking, sakata2019crownn, zhang2014spectral, shah2020permutation, chakraborty2024maxmin, siththaranjan2023distributional}, and/or require that every human label every query~\cite{parisi2014ranking, chhan2024crowd}. Important open problems in learning from multi-user feedback include, but are not limited to, enabling more and differing interaction modalities across the crowd, integrating real-time human feedback (rather than assuming all human data is available upfront), handling limited or intermittent availability of various humans, determining when (and how) to learn from potentially conflicting feedback across the crowd, effectively propagating critical information from minority crowd members, and accounting for the humans' various strengths, weaknesses, and biases.



\subsection{Human-AI Interaction Methods in which Humans Interact at Multiple Levels in a Hierarchy}\label{ssec:hierarchy}
AI systems for C2 processes will need to integrate and synchronize multiple Army echelons' actions and effects. We propose to leverage hierarchical AI methods for learning and planning, which not only have the potential to learn faster than flat architectures \cite{dayan1992feudal,parr1997reinforcement,pashevich2018modulated,dietterich2000hierarchical}, but also are built to decompose tasks for multiple levels of planning and execution \cite{kulkarni2016hierarchical,vezhnevets2017feudal,sukhbaatar2018learning,frans2017meta,bacon2017option,tessler2017deep,nachum2018data,pashevich2018modulated,brohan2023can,driess2023palm,wang2023voyager}.
Algorithms for human-AI interaction can especially benefit from a hierarchical structure, since this aligns with how humans naturally process complex tasks, breaking them down into simpler, manageable subtasks \cite{gebhardt2020hierarchical}. 
We expect AI systems that can efficiently decompose tasks into a hierarchical structure to more effectively learn from humans and to decompose high-level goals into low-level actions. Such methods hold the potential to enable future AI systems for cross-echelon C2 that function across the C2 operations process.

Further research is needed to develop hierarchical learning and planning methods applicable to complex and dynamic environments, in which humans interact at multiple levels in a hierarchy.
While existing hierarchical learning methods have shown promise, these methods typically only study two-level hierarchies, which are much simpler than hierarchies in real-world applications.
In addition, while some works study human feedback in hierarchical learning settings, e.g., via language and preferences~\cite{brohan2023can,driess2023palm,wang2023voyager,pinsler2018sample,bougie2022hierarchical,gehring2021hierarchical,wang2022skill}, the human feedback is only given at a single level in the hierarchy. Further work should also develop methods to facilitate proper synergy and communication between networks of humans and AI agents organized in multiple hierarchical levels~\cite{friston2024designing}.

\subsection{Human-AI Interaction Techniques Robust to Limited Communication and Imperfect Information Scenarios}
Previous research has proposed methods for multi-agent planning with limited inter-agent communication~\cite{heredia2019distributed, zhang2021decentralized, kim2020communication, kondo2023robust}; however, these methods consider information sharing in constrained forms, e.g., agents can only share specific predictions~\cite{kim2020communication, zhang2021decentralized, wai2018multi}, the agents' ability to communicate is determined solely based on geographic proximity~\cite{heredia2019distributed, wai2018multi}, or information transmission is subject to time delays~\cite{kondo2023robust}.
Further research is needed to develop algorithms that are robust to more sophisticated constraints on communication, analogous to DDIL conditions encountered in warfare. Another open problem involves modeling the intentions and future actions of other agents in complex and dynamic situations. In contrast, existing works~\cite{kim2020communication, meta2022human, raileanu2018modeling, wu2021too} do not robustly operate under partial observability or reliably identify decision-making factors or model-changing behaviors of other agents~\cite{albrecht2018autonomous}.

The proposed research will enable AI systems for C2 that help integrate and synchronize dispersed forces and MDO effects across dispersed command post nodes to operate under DDIL conditions, as well as with the incomplete knowledge of other entities' plans that might arise under DDIL conditions.
The problem of maximizing synergy among cooperating AI agents~\cite{friston2024designing} (see Section~\ref{ssec:multi_human_AI}) becomes even more critical under DDIL conditions, as agents observe differing information and experience hindered communication capacity.
Estimating teammates' actions under DDIL is challenging with or without AI; however, we contend that AI has the potential to improve plan intricacy and to determine plan feasibility.
Further research should develop methods that model the intentions, beliefs, capabilities, and future actions of other agents involved in a complex and dynamic scenario. Techniques such as active sensing could identify key pieces of information to send or observe to reduce uncertainty in the future actions distribution.


\subsection{Human-AI Interaction Techniques for Leveraging Adapting Databases of Human-Generated Data} Recent advances in natural language processing~\cite{vaswani2017attention,kenton2019bert,raffel2020exploring,brown2020language} have demonstrated how machine learning algorithms can learn from widely available human-generated text corpora. People can further specialize these models to solve desired language-based tasks via \textit{in-context learning}~\cite{scao2021many,min2022rethinking,xie2021explanation}, in which users simply adapt models via short snippets of text or information (\emph{prompts}), as opposed to training them from scratch with task-dependent text corpora~\cite{luketina2019survey}.
These models can be leveraged to generate plans~\cite{goecks2023disasterresponsegpt}, to reason about how to complete an objective in terms of subtasks~\cite{brohan2023can,driess2023palm}, or to propose new skills or behaviors to maximize a language-defined objective~\cite{wang2023voyager,du2023guiding}.
However, there remain a number of open research questions, including how to  
provide tailored information to humans based on their roles, their functions, or the context in which they are performing.
Another open question involves how to design AI agents that quickly adapt to new information when learning from large corpora of human-generated data, particularly when these databases expand over time to include new human data with multiple potentially-shifting human mindsets and biases. Such methods will allow C2 planning algorithms to leverage large databases of multi-modal data---from annotated maps to recordings of COA development sessions to written after action reviews---and to adapt quickly as these databases grow, e.g., maintaining robustness following personnel rotation.

\section{Effective and Resilient Human-AI Teams for Complex, Dynamic Tasks}
The seamless interaction of human-AI partnerships in any complex, dynamic environment requires understanding how to create and optimize resilient teams. The second research area focuses on the gaps related to understanding how to define the roles of humans and AI and the configurations of human-machine teams, selecting and training optimal human partners, and facilitating trust and communication across human-AI teams. By solving open research problems related to this focus area, SIML-enabled systems can be better designed for and integrated with C2 personnel performing future C2 processes.

\subsection{Defining Resilient Human-AI Roles and Configurations}\label{ssec:roles_configs} The emergence of SIML tools will shift roles of C2 personnel, and novel team configurations must be determined to combine human and AI roles together to efficiently and effectively accomplish C2 tasks. Roles and configurations within human teams are typically allocated based on the knowledge, skills, and abilities of each team member, but SIML tools enable non-traditional, flexible team roles and configurations \cite{ramchurn2021trustworthy}. Machine intelligence will be better suited for some roles and tasks that are, for example, data processing heavy or repetitive, freeing up humans for new roles, for instance, ensuring the reliability of data and recommendations from AI systems. Within the human-AI team, each partner should be able to detect limitations in the system and influence tasking to overcome shortcomings in the system by changing the configuration or roles, e.g, via via task reallocation, which can be accomplished through creating team design patterns (TDPs) \cite{schadd2022afraid}. TDPs are a promising approach to create efficient human-machine teaming workflows, and could be used to modernize processes and procedures of functional and integrating cell members within a headquarters staff. TDPs provide generic reusable behaviors across team members for supporting effective and resilient teamwork that allow for a range of combinations of human-machine partnerships based on task and type of work (i.e., cognitive vs. physical) \cite{van2018developing,van2019team}. Since a TDP breaks down a task into sub-processes based on capabilities of human or AI systems~\cite{schulte2016design}, this approach provides a design solution to address meaningful human control or ensure humans are able to make informed, timely decisions over AI systems to achieve (or prevent) a given outcome \cite{van2021moral}.

The structure of a human-AI team should be dictated by the types of human-AI interactions leveraged, such as mediating AI decision recommendations or providing feedback on RL policies \cite{ramchurn2021trustworthy}. Emerging research focused on understanding consistent roles within human-AI teams suggests supervising, collaborating, and overseeing as potential human roles \cite{onnasch2021taxonomy, siemon2022elaborating}. An SIML system, for example, focused on creating an accurate Common Operating Picture (COP) may have data wranglers and human deception checkers to validate incoming data from sensors and sources. Building on this, information from SIML systems should convey tailored information depending on human roles, function, and context. C2 personnel focused on targeting, for example, could distill or represent information within the AI system that allows for optimizing task roles, skill knowledge, and context. Role testing and evaluation will be critical steps in defining human and AI roles, for instance facilitating improvements in aspects of human workload or exploitation of the AI. For example, Ramchurn et al.~\cite{ramchurn2016disaster} found that a human-AI partnership performed best on a team scheduling task when a human role required mediating between human and AI partners, allowing the AI to overcome prediction constraints and unexpected outcomes.

Open research problems in this area include understanding how human roles in a multi-human and multi-AI team change as the AI systems become more capable, how roles and compositions need to change as the complexity, time-sensitivity, and uncertainty of the problem space increase, and how to create TDPs efficiently for C2 processes. Operational requirements will drive development of SIML tools and interactions, but addressing these problems will allow for a range of potential available resources in human and AI partners. Optimization of human and AI roles along with their configuration within the team must stem from systematic testing and evaluation based on comparing overall performance outcomes to determine how to create teams that scale and apply across a multitude of operational settings. Development of C2 personnel and staff TDP libraries, as part of Army knowledge management, could serve as doctrinal templates that help form foundational planning standard procedures and guide collective C2 work \cite{van2018developing}.

\subsection{Effective Human-AI Partnerships through Selecting and Training Optimal Human Partners} A budding area of research seeks to understand characteristics of human partners that result in better human-AI integration. Specifically, how do cognitive abilities, personality traits, and attitudes influence interactions with SIML systems, and what makes some people optimal human partners for a given AI system? A theoretical framework proposed by Hoff et al.~\cite{hoff2015trust} suggests that dispositional characteristics, such as personality traits, gender, and age, along with personal experiences and context of use, can greatly influence dynamics of calibrated trust and subsequent interactions with AI or autonomous systems. Achieving calibrated trust results in minimizing distrust that can lead to errors, disuse, and over-reliance, while reducing over-trust that can lead to miscalculations in risk~\cite{wickens2015complacency,lee2004trust}. Training programs aimed at calibrating trust in human-machine teams have proven effective, since trust can evolve over time ~\cite{johnson2023impact}.

A key aspect of human-AI interactions is the formation of a shared mental model, i.e., a shared understanding of the team's task, each team member's role in it, and their respective capabilities, which allows for the predictions of needs and behaviors of other team members \cite{andrews2023role}. This suggests that traits such as empathy, theory of mind, or emotional intelligence might allow a human to be flexible in their mental model of the AI agent. 
Selection and understanding of these traits in human partners could then advantageously remedy inherent AI bias, provide over-correction in training data, or result in more flexible mental models that are more resilient to changes in AI systems. Emerging work with human-AI teams in code translation tasks suggests human teammates can overcome limitations in AI capabilities by successfully performing tasks with imperfect AI \cite{weisz2021perfection}. Similarly, reasoning over quantified uncertainty is a human skill that can be improved, in some contexts, with deliberate practice~\cite{kusumastuti2022practice, kay2016ish}.
SIML systems will differ in their sophistication and ability to be used effectively amongst the general population. Research is needed to identify characteristics of optimal human partners and human-AI compatibility; such work will aid in alignment of roles and configurations to result in the best performance outcomes and robust, ethical human-AI partnerships~\cite{mcneese2021my}. While work is ongoing to define and model AI technological fluency \cite{Campbell2023}, additional predictors of successful human-AI partnering might go beyond more general use of AI. Compositionality is a concept related to fluid intelligence that explains cognitive flexibility as the ability to tackle complex tasks by combining simpler cognitive operations. Seeking this ability in both human and AI teammates could serve as a foundation for successful human-AI partnerships~\cite{Duncan2017}. Given that AI capabilities and vulnerabilities change rapidly, and these changes are likely to accelerate, human ability to promptly identify and adapt to changing operational characteristics may be paramount \cite{pollard2022prepare}. 

Open questions in this area include better understanding what human characteristics are relevant to successful interaction with SIML systems; how best to develop a workforce with those characteristics, be it through selection, education, training, or other approaches; and how to make sure humans are able to keep up with the rapidly changing AI landscape. Addressing these questions would potentially enable a greater population to effectively modify and guide AI systems, and ensure humans thrive as members of human-AI teams.

\subsection{Communication through Explainable AI and Shared Mental Models}

Within teaming situations, communication~\cite{mcneese2021my} and development of a shared mental model are important for humans to accurately predict behaviors of both AI~\cite{andrews2023role} and other humans.
Just as a human teammate might be required to provide supporting reasoning and evidence for a recommendation, AI will need to provide explanations to ensure a shared understanding of the costs and benefits of a particular recommendation. 
While AI decision-making methods such as RL are capable of rapidly exploring numerous decision sequences through policies that map observations to actions, these policies often result in black-box functions that are challenging for humans to explain or justify due to their unintuitive nature~\cite{milani2022survey}. 

Explainable systems~\cite{confalonieri2021historical} may help humans to build calibrated trust~\cite{cha2019visualization, zhang2020effect} in the AI system, but explanations that are not well suited to the task or context might lead to miscalibrated trust~\cite{bansal2021does, vasconcelos2023explanations}. Miller~\cite{miller2019explanation} argues that AI must leverage humans' mental models of the world to generate effective explanations. Consistent with this idea, Neerincx and colleagues \cite{neerincx2018using} propose an explainable AI approach that is transferable across domains and focuses on explaining an agent's behavior in terms of its data-driven (perceptual; e.g., contrastive explanations and confidence metrics) and model-based (cognitive, e.g., beliefs and goals) processes while considering the use context and human user capabilities in explanations. This is consistent with another approach to increasing trust between humans and AI through communication by accounting for uncertainty of AI system recommendations and solutions via AI-provided confidence ratings~\cite{zhang2020effect}. Approaches for explainability can be integrated and executed through use of TDPs \cite{van2021moral, miller2019explanation}, which may shape the human partner's mental model of the AI \cite{schadd2022afraid, miller2019explanation}. 
But, formation and shaping of the AI mental model is not so easy. One approach is to create an AI compatible representation of human behavior or knowledge and for the AI to validate its understanding back to the human user. Recently, this approach was successfully applied to mission planning, specifically understanding commander's intent, to accelerate planning and decision making processes~\cite{schadd2022machine}. 


Open problems in this area include creating effective methods for increasing communication and shaping shared mental models across human-AI partnerships and that apply to complex, high-dimensional, and dynamic planning tasks that will be found in future C2 operational environments. Research is needed to extend 
policy-based human-AI interaction approaches to predict how actions will unfold~\cite{reddy2020learning, liu2023efficient} while accounting for uncertainty. 
Recent progress has been made in conveying uncertainty in relatively simple contexts~\cite{padilla2021review, kale2018hypothetical}, 
but these methods become computationally intractable under high complexity~\cite{ghahramani2015probabilistic} or account only for model uncertainty~\cite{gal2016dropout}, and thus modeling uncertainty remains an open problem. 
Further research should also develop methods that convert policies to explainable plans, e.g., via language, annotated maps, or trees of potential outcomes. Such approaches could leverage generative AI tools such as LLMs and image generation models~\cite{openai2022introducing, kenton2019bert, bubeck2023sparks, marcus2022very} and combine machine learning methods for sequential planning (e.g., RL) with generative AI~\cite{meta2022human} to convey information to human users that improves a shared mental model. Related, a remaining challenge is measuring and eliciting shared mental models in experimentation to optimize human-AI partnerships~\cite{andrews2023role}. 

\section{Flexible and Scalable Decision-Making with Interactive Machine Learning} 

To deploy human-AI interaction systems not only in simulations and simplified exercises, but also in real-world situations, AI algorithms will need to scale flexibly and robustly in response to a variety of dynamic factors. Our third research focus area considers flexibility across scales, particularly in regard to decision-making timescales, the numbers of interacting human and AI agents, the type of decision-making hierarchy, and the problem sphere or scope. Critically, AI systems must not only scale, e.g., as the number of humans increases, but must also flexibly adapt as scales vary bidirectionally in real-time, e.g., if suddenly only two humans are available to interact with the AI system, and as a result, certain expertise is missing.
We discuss open research opportunities in each of these areas. Addressing these gaps will be key to enabling scalable human-AI partnerships that can learn and plan in complex and dynamic environments, as are prevalent in C2.

\subsection{Temporal Scalability}
Algorithms for human-AI interaction must function smoothly across both longer and shorter decision-making timescales. This is important for C2 applications, since depending on circumstances, the available decision-making time could vary from minutes or hours to days, while possibly respecting further constraints such as the 1/3-2/3 rule.\footnote{From U.S. Army Doctrine Publication 5.0, ``The Operations Process'': ``Commanders follow the `one-third, two-thirds rule' as a guide to allocate time available. They use one-third of the time available before execution for their own planning and allocate the remaining two-thirds of the time available before execution to their subordinates for planning and preparation''~\cite{army2019ADP5-0}.}
Algorithms should adapt their behavior given the available time, e.g., by intelligently trading off between factors such as solution optimality, the number of explored decision alternatives, and a temporal budget. Furthermore, temporal budgets could in turn constrain resources such as computational time and human time spent interacting with the AI. Such trade-offs could affect AI behavior in many ways, for instance by informing the number of AI-generated alternatives (e.g., COAs) shown to C2 personnel, the number of simulations or wargames in which these alternatives are evaluated (e.g., COA analysis), the choice of \textit{which} alternatives to present to personnel (e.g., COAs with more predictable outcomes vs. high-risk, high-reward alternatives), and the chosen modalities of human-AI interaction (e.g., querying personnel for detailed feedback vs. a simple COA selection among presented alternatives).

While some existing AI decision-making methods consider a query budget that limits algorithm exploration~\cite{taylor2014reinforcement, samadi2015askworld, ilhan2019teaching}, consider the time required to process input features~\cite{xu2012greedy}, trade off between speed and accuracy 
~\cite{basu2019pareto}, or use a temporal budget to inform the fidelity at which to run simulations (i.e., balancing between simulation speed and accuracy)~\cite{song2019general}, these methods apply to limited settings that are significantly simpler than C2 scenarios. Furthermore, they often utilize no human interaction~\cite{taylor2014reinforcement, song2019general, ilhan2019teaching, xu2012greedy}, or at most a single human interaction modality~\cite{samadi2015askworld, basu2019pareto}, and often only employ the budget constraint to determine when to halt algorithm execution~\cite{taylor2014reinforcement, samadi2015askworld, ilhan2019teaching}. Further research should develop human-AI interaction methods that can flexibly scale to a range of temporal constraints, for instance determining the best allocation of human and computational resources and the optimal human interaction approach for the available time. Future research should also consider how multiple decision-making AI agents operating at different timescales can best synergize with one another, as well as with humans.

\subsection{Human-AI Interaction Scalability}
Human-AI integration techniques must also scale across changing numbers of human and AI agents in the system, including as availability of human or AI agent resources increases or decreases in real-time. Algorithms should be capable of integrating feedback from many people, while retaining robustness when only a few people---or even just a single person---are available to interact or provide feedback. 
Future work should develop decision-making methods that not only can accept feedback from multiple humans (Section~\ref{ssec:multi_human_AI}) and are cognizant of these humans' diverse roles and expertise (Section~\ref{ssec:roles_configs}), but also that are flexible given shifts in the numbers and expertise of accessible humans, including in dynamic scenarios, to maximize human resources while filling in when people become unavailable. In C2 scenarios, such approaches would integrate feedback from many battlefield assets and C2 personnel, while remaining robust when only few C2 personnel are present, seamlessly adapting based on the specific roles and functions of available staff.

Decision-making algorithms must also scale to robustly direct either many or few AI agents. Existing algorithms for multi-agent learning and planning often become computationally intractable as the number of AI agents increases or cannot adapt if AI agents are dynamically added to or removed from the system~\cite{reifsteck2019epistemic, kaduri2020algorithm, kasaura2022prioritized, wong2023deep}.
We envision improved algorithms for human-AI integration that scale to plan over many (hundreds or more) forces and assets on the battlefield, while also reliably generating plans when comparatively few assets are present, or when asset availability shifts (e.g., decreases) in real-time. Ensuring synergy between many AI algorithms that receive input from many humans is an open research area; for instance, Friston et al.~\cite{friston2024designing} suggest research avenues for coordinating among a network of many interacting AI systems to achieve collective intelligence.

\subsection{Hierarchical Scalability}
To tackle sophisticated problem scenarios such as arise in C2, human-AI partnerships will need to be organized hierarchically, with structured human-AI interactions occurring at multiple hierarchical levels (see Section~\ref{ssec:hierarchy}). Furthermore, however, these algorithms will need to flexibly operate across a range of potential hierarchical structures, switching between them as needed, e.g., to rapidly adapt to changing battlefield conditions.
Further research should develop hierarchical algorithms for human-AI interaction that can scale across many hierarchical organizations, e.g., with varying numbers of hierarchical levels and extending to tree-like hierarchical structures. These algorithms should also flexibly adapt in real-time if hierarchies change based on situational needs.
In the C2 space, such algorithms will promote robust human-AI interaction across various multi-echelon organizational structures, assisting each echelon and unit individually while maintaining unity of effort. 
They could also help to coordinate among hierarchies within different countries with different equipment, data, etc. in a single joint mission command situation, e.g., directly facilitating NATO interoperability~\cite{derleth2015enhancing}.

\subsection{Problem Sphere Scalability}
In complex domains such as C2, learning and planning algorithms should identify the sphere (or scope) of the problem that is most crucial for effective decision-making, and then autonomously adapt decision-making to the key factors within this sphere. For instance, such a capability could help to focus algorithmic planning for C2 on the most relevant geographic regions; perhaps, for the current operation, activity in the east must be estimated more accurately than activity in the west, while for a different operation, activity must be precisely estimated over a larger region of the battlefield. Meanwhile, AI planning should also scale its focus to concentrate on the most relevant warfighting functions; for example, perhaps for the current mission, fires must be estimated more accurately than sustainment. AI algorithms may need to adapt their focus in real-time to changing regions of interest (whether geographical, operational, etc.), while different echelons or dispersed forces may have differing modeling priorities or require differing levels of abstraction. Algorithms that leverage previous experience and human interaction to scale their computational focus to a problem's most critical aspects would facilitate improved utilization of constrained temporal and computational resources. While multi-resolution modeling can enable AI to learn from data at different abstraction or granularity levels~\cite{mangalraj2020review, zheng2021multi, wang20213m}, future research in 
human-AI integration 
should expand on these techniques to promote both scalability and adaptability across a broad range of problem spheres, as may arise in complex military situations. 

\section{Discussion}
An SIML-enabled C2 system has the potential to revolutionize C2 personnel effectiveness and efficiency to achieve decision advantage over peer and near-peer adversaries. Yet, additional research is needed to develop the SIML science and technology capabilities necessary for 
approaches to be practical and robust in the C2 domain. Here, we proposed and described three critical research focus areas for enabling SIML systems for C2. First, successful SIML relies on leveraging human-AI interactions 
for complex planning,
e.g., for multiple humans and multiple agents to collaborate together via a variety of interaction modalities and across hierarchical levels. We expect research in this area to develop over the next decade.
Second, while the current MDMP is a fundamentally human endeavor, an SIML-enabled MDMP must reconsider who and what AI systems comprise a given team, along with optimal team roles and configurations given available C2 personnel. Future research should aim to better understand how to optimize team performance through selection, training, and role assignment. We expect this newly emerging field of human-AI teaming performance and optimization to develop over the next decade, whereas the application of the science may begin to emerge as early as 2040. Third, advances in human-AI integration and resilient teaming will only be meaningful for C2 situations if they are flexible and scalable to a broad range of potential operational contexts. While perhaps the most difficult to execute, this research focus area is critical to the transition of such SIML-enabled systems to real-world C2 contexts. Given the validation and testing requirements of military technology along with additional security standards and protocols, we expect to see smaller applications first appear more widely within the next 5 years, followed by more advanced SIML-enabled systems by 2040. 

Finally, practical considerations must be taken into account when deploying SIML science and technology capabilities at the tactical edge. In line with industry trends of incorporating AI technologies into smartphones and personal computers, though in a miniaturized version, we foresee that the computing and power requirements for future AI technologies will be met for warfighters at the tactical edge. This vision aligns with ongoing advancements in AI, suggesting that these critical capabilities will be accessible to future C2 personnel potentially as early as 2040. In addition, there are many open considerations in the implementation of SIML-systems for C2 operational environments, such as the choice between pre-learned policies versus online learning or the type of architecture for a given system. Perhaps most important are practical considerations that impact aspects of interoperability between NATO and Ally forces. Interoperability allows different Allies to operate together in a coherent and efficient manner to achieve tactical, operational, and strategic objectives through common communication, standards, procedures, equipment, and doctrine. Related decisions regarding data and model management are critical, encompassing how data is stored, accessed, and utilized for learning, as well as how models are maintained, updated, and managed over time. As time evolves, NATO and other multi-national agreements, such as STANAG, a NATO Standardization agreement, will need to adapt to emerging technology, such as SIML-enabled systems, to modernize interoperability objectives \cite{derleth2015enhancing}. 

\section*{Acknowledgements}
The authors thank Mr. Lee Vinson from the Apex Analytics Group and the Mission Command Battle Lab at Fort Leavenworth, Kansas, for their insightful feedback and discussion when refining the concepts proposed in this paper.

\bibliographystyle{ieeetr}
\bibliography{References_condensed}

\end{document}